\documentclass{article}
\usepackage{ijcai22}

\usepackage{times}
\usepackage{soul}
\usepackage{url}
\usepackage[hidelinks]{hyperref}
\usepackage[utf8]{inputenc}
\usepackage[small]{caption}
\usepackage{graphicx}
\usepackage{amsmath}
\usepackage{amsthm}
\usepackage{booktabs}
\usepackage{algorithm}
\usepackage{epsfig}
\usepackage{amsmath}
\usepackage{amssymb}
\usepackage{amsfonts}       
\usepackage{nicefrac}       
\usepackage{microtype}      

\usepackage{color}
\usepackage{array}
\usepackage{comment}
\usepackage{cases}
\usepackage{algpseudocode}
\usepackage{multirow}
\usepackage{textcomp}
\usepackage{subfigure}
\usepackage{booktabs}
\usepackage[export]{adjustbox}
\usepackage{bbm}
\usepackage[switch]{lineno}

\usepackage{multirow}
\usepackage{pifont}
\usepackage{makecell}
\usepackage{enumitem}
\usepackage{colortbl}

\definecolor{mygray}{gray}{.95}
\definecolor{mywhite}{gray}{0.99}

\newcommand{\cmark}{\ding{51}}%
\newcommand{\xmark}{\ding{55}}%

\newcommand{\eg}{\textit{e.g.}}
\newcommand{\ie}{\textit{i.e.}}
\newcommand{\vs}{\textit{v.s.}}

\usepackage{algorithm}
\usepackage{algorithmicx}
\usepackage{cases}
\usepackage{algpseudocode}
\usepackage{multirow}
\usepackage{textcomp}
\usepackage{subfigure}
\usepackage{booktabs}

\usepackage[switch]{lineno}
\usepackage{booktabs}
\usepackage[export]{adjustbox}
\usepackage{bbm}

\usepackage{listings}
\usepackage[symbol]{footmisc}

\usepackage{dblfloatfix}
\usepackage{transparent}
\usepackage{etoolbox}
\makeatletter
\AfterEndEnvironment{algorithm}{\let\@algcomment\relax}
\AtEndEnvironment{algorithm}{\kern2pt\hrule\relax\vskip3pt\@algcomment}
\let\@algcomment\relax
\newcommand\algcomment[1]{\def\@algcomment{\footnotesize#1}}
\renewcommand\fs@ruled{\def\@fs@cfont{\bfseries}\let\@fs@capt\floatc@ruled
  \def\@fs@pre{\hrule height.8pt depth0pt \kern2pt}%
  \def\@fs@post{}%
  \def\@fs@mid{\kern2pt\hrule\kern2pt}%
  \let\@fs@iftopcapt\iftrue}
\makeatother

\title{Towards \emph{Universal} Backward-Compatible Representation Learning}

\iftrue
\author{
Binjie Zhang$^{1,2}$\footnotemark[3]
\and
Yixiao Ge$^2$\footnotemark[2]\and
Yantao Shen$^{4}$\footnotemark[3]\and
Shupeng Su$^2$\and
Fanzi Wu$^{4}$\and\\
Chun Yuan$^1$\footnotemark[2]\and
Xuyuan Xu$^3$\and
Yexin Wang$^3$\and
Ying Shan$^2$
\affiliations
$^1$Tsinghua University \quad 
$^2$ARC Lab, Tencent PCG \\
$^3$AI Technology Center of Tencent Video \quad 
$^4$AWS/Amazon AI\\
\small{
\texttt{\{zbj19@mails,yuanc@sz\}.tsinghua.edu.cn} \quad 
\texttt{\{yixiaoge,yingsshan\}@tencent.com}
}
}
\fi

\begin{document}

\maketitle

\renewcommand{\thefootnote}{\fnsymbol{footnote}}
\footnotetext[2]{Corresponding authors.} 
\footnotetext[3]{Work done when Binjie and Yantao are at ARC Lab.}

\begin{abstract}
    Conventional model upgrades for visual search systems require offline refreshment of gallery features by feeding gallery images into new models (dubbed as ``backfill''), which is time-consuming and expensive, especially in large-scale applications.
    The task of backward-compatible representation learning \cite{Shen_2020_CVPR} is therefore introduced to support backfill-free model upgrades, where the new query features are interoperable with the old gallery features.
    Despite the success, previous works only investigated a close-set training scenario (\textit{i.e.}, the new training set shares the same classes as the old one), and are limited by more realistic and challenging open-set scenarios. 
   To this end, we first introduce a new problem of universal backward-compatible representation learning, covering all possible data split in model upgrades.
   We further propose a simple yet effective method, dubbed as \textbf{Uni}versal \textbf{B}ackward-\textbf{C}ompatible \textbf{T}raining (\textit{Uni}BCT) with a novel structural prototype refinement algorithm, to learn compatible representations in all kinds of model upgrading benchmarks in a unified manner.
   Comprehensive experiments on the large-scale face recognition datasets MS1Mv3 and IJB-C fully demonstrate the effectiveness of our method.

\end{abstract}
   
   \begin{figure}[ht]
   \begin{center}
   \includegraphics[scale=0.4]{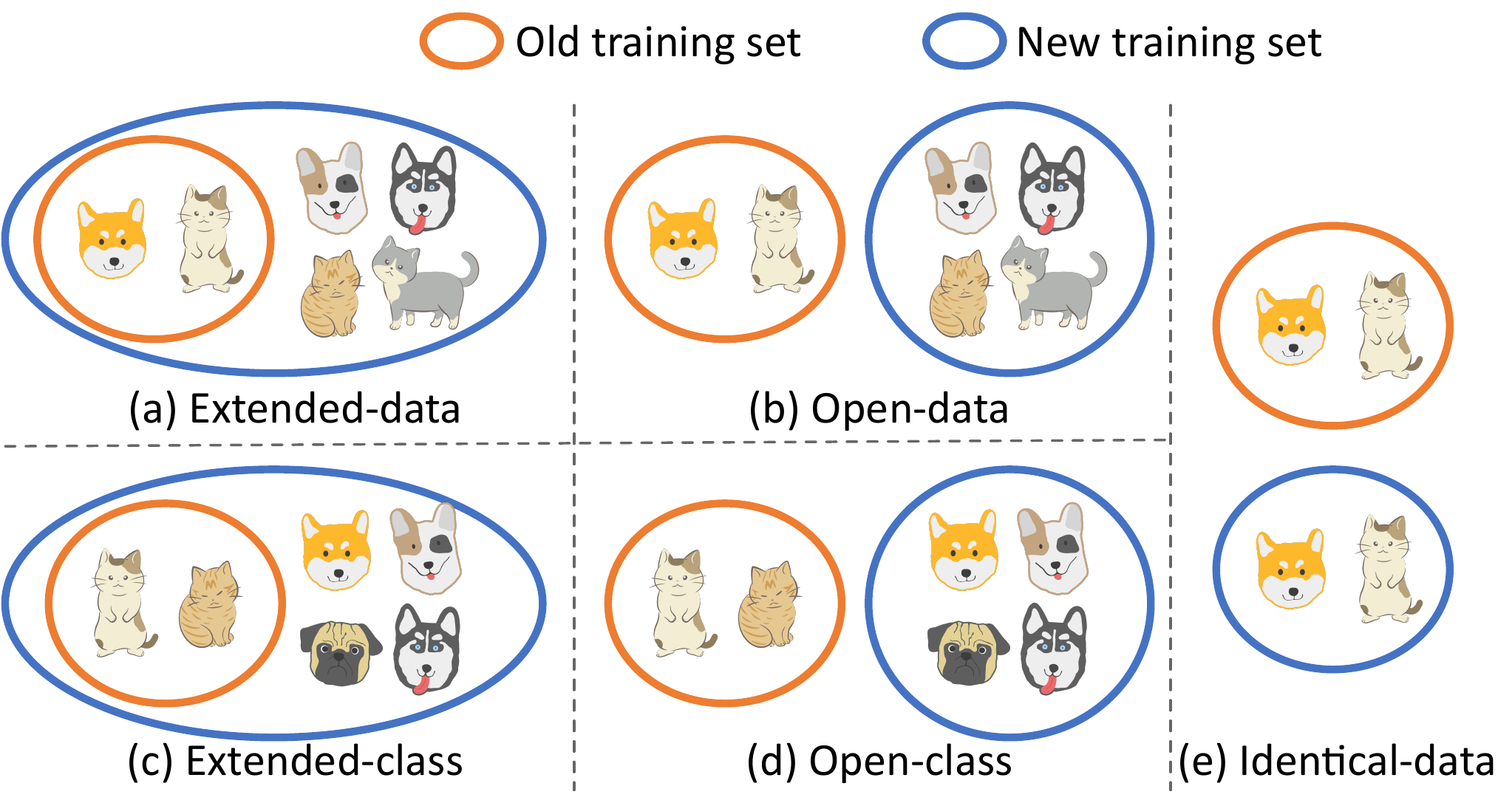}
   \end{center}
   \vspace{-10pt}
   \caption{Illustration of different training data distributions for universal backward-compatible training. According to the data and category differences between old and new training sets, we summarize the data splits into five types from (a) to (e), covering most of the compatible training scenarios for backfill-free model upgrades.}
   \label{fig:teasor}
  \vspace{-10pt}
   \end{figure}
   
   \section{Introduction} 
   \label{sec:introduction}
   Deep learning-based methods~\cite{ghifary2016deep,he2016deep,li2017learning} have achieved great success in visual search tasks, such as face recognition~\cite{liu2017sphereface,yang2017neural,wang2017normface,wang2018cosface,deng2019arcface,zhang2020rbf} and landmark retrieval~\cite{oxford:philbin2007object,paris:philbin2008lost,gldv2:weyand2020google,roxford:radenovic2018revisiting,ge2020self}. 

The task of visual search requires to retrieve the same objects' images from a large-scale database (dubbed as \textit{gallery}), given an image of interest (dubbed as \textit{query}).
The process of offline ``backfilling''\footnote{Since the upgraded (new) model is not directly comparable with the old gallery features, the gallery needs to be re-extracted via feeding all the raw images into the new model.} the gallery is always necessary for conventional model upgrades in retrieval systems, which is computationally expensive and time-consuming. Moreover, it is infeasible when the raw images are inaccessible due to privacy issues or storage limitations.

    Thanks to the introduction of backward-compatible representation learning \cite{Shen_2020_CVPR,zhang2021hot}, new models that are trained with compatibility constraints can be immediately deployed in a backfill-free manner, where the encoded new features for queries are interoperable with the old gallery features. 
    The follow-up works make efforts to further improve the feature compatibility by designing advanced training constraints~\cite{budnik2020asymmetric,meng2021learning} or transformation architectures~\cite{wang2020unified}.
   Positive as the results are, they only focused on a single close-set model upgrading scenario (dubbed as extended-data in Figure~\ref{fig:teasor} (a)), where the new training data share the identical class set as the old one.
   It is notable that the data split for model upgrades in real-world applications is complex and unpredictable, that is, both close-set and open-set scenarios should be considered.
   Existing methods \cite{Shen_2020_CVPR,wang2020unified,budnik2020asymmetric,meng2021learning} did not investigate the open-set data split and are even inapplicable in such a scenario. 
   
   Towards this end,
   we for the first introduce the task of \textit{universal} backward-compatible representation learning, where five kinds of data split covering both close-set and open-set scenarios are considered, as demonstrated in Figure~\ref{fig:teasor}.
   The open-set data split (including extended-class, open-data and open-class) poses a great challenge for learning compatible representations due to the potential domain gaps among different data and categories.
   
   To tackle the challenge,
   we introduce a simple yet effective method, namely \textbf{Uni}versal \textbf{B}ackward-\textbf{C}ompatible \textbf{T}raining (\textbf{\textit{Uni}BCT}), to encode compatible representations in all kinds of data splits in a unified manner.
   Specifically, inspired by \cite{Shen_2020_CVPR}, we utilize the old classifier (in the form of a fully-connected layer) to provide valuable supervision from the old latent space, \textit{i.e.}, enforcing the new features to be closer to their corresponding old class centers.
   As for the novel categories in the open-class and extended-class scenarios,
   we extract the features of the new categories' images and leverage their class centroids to construct pseudo prototypes.
   Due to the category gaps~\cite{you2019universal}, the pseudo prototype inevitably carries some noise that may affect the representation learning of backward compatibility.
   Therefore we propose to improve the class centroids of the pseudo prototype via a novel structural prototype refinement algorithm, \textit{i.e.}, the ``old'' features of the new classes' images are refined by propagating their neighbors' knowledge via a fully-connected graph.
   The graph works under the assumption that visually similar images (measured by the new model which has the stronger capability) should have close-by old features.

   In a nutshell, our contributions are three-fold.
   (1) We introduce a new task, namely universal backward-compatible representation learning, which aims at investigating all possible data splits in practical model upgrading scenarios.
   (2) We propose a novel method, dubbed as universal backward-compatible training (\textit{Uni}BCT), to tackle the challenge of different kinds of data splits in a unified manner. Our method is simple yet effective to refine the noisy pseudo prototype and improve the feature compatibility on both close-set and open-set scenarios.
   (3) We conduct comprehensive experiments on the large-scale face recognition datasets MS1Mv3
 \cite{deng2019arcface} under five different model upgrading benchmarks , and investigate different compatibility constraints via the evaluations on IJB-C \cite{ijbc_maze2018iarpa}. Our \textit{Uni}BCT consistently outperforms the baseline and other advanced regularizations, fully indicating the effectiveness of our method.
   
   \begin{figure*}[t]
      \begin{center}
      \includegraphics[width=1\linewidth]{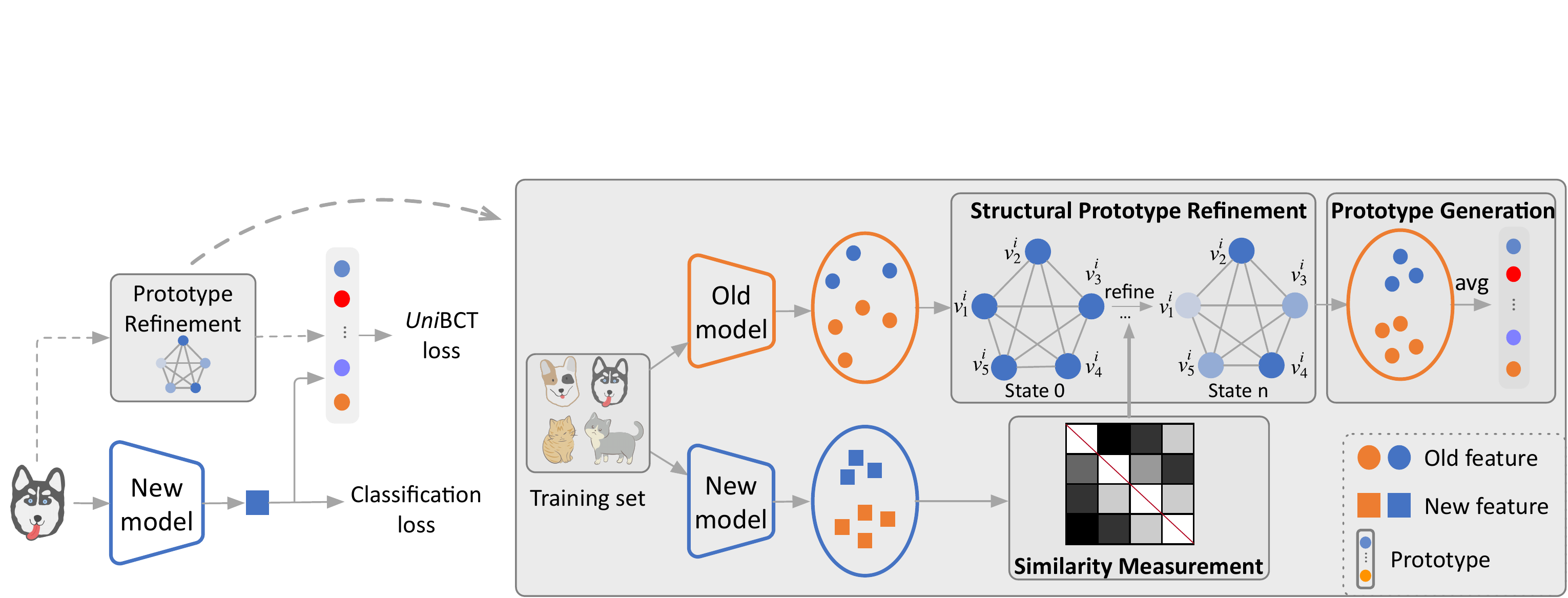}
      \end{center}
      \vspace{-8pt}
      \caption{Pipeline of our Universal Backward-Compatible Training (\textit{Uni}BCT). The new model is supervised by a classification loss to learn discriminative features, and an additional universal backward-compatible loss to make sure the new features be interchangeable with the old ones. To alleviate the negative effects of data and category gaps between old and new training data in open-set model upgrading scenarios, we introduce a novel module named structural prototype refinement. It improves the old feature quality by propagating their neighbors' knowledge via a fully-connected graph. 
      Note that during the training process, the pseudo prototypes will be not updated by the loss backpropagation.
      }
      \label{fig:method}
   \end{figure*}
   
\section{Related Work} 
   \label{sec:related_work}
   
   \paragraph{Backward-Compatible Learning}
   aims to make new features and the old ones interoperable and realize backfill-free model upgrades.
   \cite{Shen_2020_CVPR} first formulated the problem by deriving influence loss from an empirical criterion, and solved it by utilizing the old classifier to regularize the optimization process.
   \cite{wang2020unified} proposed Residual Bottleneck Transformation (RBT) blocks for feature embedding transferring. 
   In~\cite{budnik2020asymmetric}, authors investigated the problem of the asymmetric test, where the database images are encoded by a teacher model and {query} images are encoded by a student model. A pair-based metric for instance-level image retrieval was proposed to achieve the goal. \cite{meng2021learning} extended RBT blocks and designed advanced boundary loss to obtain more compact intra-class distributions.
   Though the above works could properly improve the compatible performance, they severely rely on the old training data or class. The open-set compatible scenarios are never investigated before.
   
   \paragraph{Universal Domain Adaptation.} While it is true that universal domain adaptation (UDA)~\cite{you2019universal,saito2020universal} and our universal backward-compatible representation learning both take data/category domain gaps between old and new training data into consideration, they have entirely different purposes. UDA focuses on transferring the model knowledge from the old domain to the new one and only requires the model to perform well on the new domain without any cross-domain operations. Universal backward-compatible learning requires the new model to encode backward-compatible features that can be directly compared with the old features.
   
\section{Universal Backward-Compatible Representation Learning} 
   \label{sec:approach}
   In this section, we first investigate the problem settings of universal backward-compatible representation learning in Sec.~\ref{subsec:problem_setting}. 
   Then  we introduce our universal backward-compatible training (\textit{Uni}BCT) method in Sec. \ref{subsec:unified_backward_compatible_training}.
   
   \subsection{Problem Settings}
   \label{subsec:problem_setting}
    Given the gallery features extracted by the old model, 
    backward-compatible representation learning task requires the trained new model to encode query features that can be directly indexed by the old gallery features.
    In real-world applications, the new training set may differ from the old one in the aspects of data or classes, raising a universal backward-compatible representation learning problem.
    
    \vspace{8pt}
    \noindent \textbf{Symbol Definition.} We denote the training set, {gallery} set, and {query} set as ${\cal D}$, ${\cal G}, {\cal Q}$. An old model $\phi_{\rm o}$ trained on old training set ${\cal D_{\rm o}}$ embeds an image $x$ to a feature vector $v_{\rm o} = \phi_{\rm o}(x)$.
    For model upgrades, a new model $\phi_{\rm n}$ trained on ${\cal D_{\rm n}}$ is obtained. The new model $\phi_{\rm n}$ embeds the image $x$ into a new feature vector $v_{\rm n}$.

    \vspace{8pt}
    \noindent \textbf{Benchmarks.} 
   Taking both close-set and open-set model upgrading scenarios into consideration,
   we discuss five kinds of dataset settings as depicted in Table~\ref{tab:compatible_scenarios}:
   (1) \textbf{Extended-data}: The old training set ${\cal {D}}_{\rm o}^{30\%-d}$ composes of 30\% images which are randomly sampled from the whole dataset, and the new training set ${\cal {D}}_{\rm n}^{100\%}$ is made up of 100\% data. The old and new training sets share the same classes.
   (2) \textbf{Open-data}: The new training data ${\cal {D}}_{\rm n}^{70\%-d}$ and the old data ${\cal {D}}_{\rm o}^{30\%-d}$ are exclusive from each other but they share the same classes.
   (3) \textbf{Extended-class}: We randomly pick 30\% classes for the old training set ${\cal {D}}_{\rm o}^{30\%-c}$ and 100\% classes for the new one ${\cal {D}}_{\rm n}^{100\%}$.
   (4) \textbf{Open-class}: Both the data and the class are different between ${\cal {D}}_{\rm o}^{30\%-c}$ and ${\cal {D}}_{\rm n}^{70\%-c}$.
   (5) \textbf{Identical-data}: The new training set ${\cal {D}}_{\rm n}^{30\%-d}$ and the old one ${\cal {D}}_{\rm o}^{30\%-d}$ are identical.

   \begin{table}[t]
      \small
      \centering
       \setlength{\tabcolsep}{4.6 pt}
      \begin{tabular}{lcccc}
         \toprule
         \multirow{2}{*}{Allocation type} & \multicolumn{2}{c}{Old train-set} &  \multicolumn{2}{c}{New train-set}\\
         &  \# images & \# classes &  \# images & \# classes \\
         \midrule
         Extended-data & 1,511,514 & 93,431 & 5,179,510 & 93,431 \\
         Open-data & 1,511,514 & 93,431 & 3,667,996 & 93,431 \\
         Extended-class & 1,549,785 & 28,029 & 5,179,510 & 93,431 \\
         Open-class & 1,549,785 & 28,029 & 3,629,725 & 65,402 \\
         Identical-data & 1,511,514 & 93,431 & 1,511,514 & 93,431 \\
         \bottomrule
      \end{tabular}
      \vspace{-5pt}
      \caption{Five different allocations for the training data, where all the images are sampled from MS1Mv3. The ``extended-data'', ``open-data'' and ``identical-data'' setups share the same old training set.}
      \label{tab:compatible_scenarios}
   \end{table}
   
   \vspace{8pt}
   \noindent \textbf{Compatibility Evaluation.} Cross-model compatibility means that the {gallery} features produced by $\phi_{\rm o}$ can be directly comparable with the {query} features extracted by $\phi_{\rm n}$.  Following~\cite{Shen_2020_CVPR}, we claim that the feature compatibility is achieved if the following empirical criterion is satisfied,
   \begin{equation} \label{eq:empirical_criterion}
       \mathcal{M} (\phi_{\rm n}, \phi_{\rm o}; \cal Q, \cal G) > \mathcal{M} (\phi_{\rm o}, \phi_{\rm o}; \cal Q, \cal G),
    \end{equation}
   where $\mathcal{M}$ is an evaluation metric for the corresponding test set.
   \textbf{Cross Test}, denoted as $\mathcal{M} (\phi_{\rm n}, \phi_{\rm o}; \cal Q, \cal G)$, is the query-to-gallery retrieval performance, where query features are extracted by new model $\phi_{\rm n}$ and gallery ones with old model $\phi_{\rm o}$. \textbf{Self Test} reflects the performance where query and gallery features are extracted by the same model (\eg, the old one). 
  
  \subsection{Universal Backward-Compatible Training}  
   \label{subsec:unified_backward_compatible_training}

   To achieve compatibility in new model training process, two universal objectives are essential: (1) obtaining discriminative feature representations for better performance, and (2) making old and new representation features interoperable. 
   The overall training objective of our universal backward-compatible training can be therefore formulated as
   \begin{equation}
   \label{eq:all_loss}
      \begin{split}
      \mathcal{L} = \mathcal{L}_{\rm cls} + \eta\mathcal{L}_{\rm uniBCT},
      \end{split}
   \end{equation}
   where $\mathcal{L}_{\rm cls}$ is the classification loss to achieve the first goal, $\mathcal{L}_{\rm uniBCT}$ is the universal backward-compatible loss to achieve the second goal and $\eta$ is the loss weight.
   
   Specifically, 
   following the state-of-the-art method in metric learning, 
   we use the form of ArcFace loss~\cite{deng2019arcface} to regularize the pretext task of classification, that is, 
   \begin{align}
       \mathcal{L}_{\rm cls}=\ell_{\rm arc}(\omega_{\rm n},\phi_{\rm n}),
   \end{align}
   where $\omega_{\rm n}$ and $\phi_{\rm n}$ denote the classifier and backbone of the new model. The formulation of ArcFace loss is
   \begin{equation}\label{eq:arcface_loss}
      \begin{split}
          &\ell_{\rm arc}(\omega,\phi)\\ 
          &= -\frac{1}{|{\cal D}_{\rm n}|}\sum_{x\in {\cal D}_{\rm n}}{ \log{ \frac{e^{s(\cos(\theta_{y}+m))}}{e^{s(\cos(\theta_{y}+m))}+\sum_{j\neq y}e^{s\cos{\theta_j}}} }}, 
      \end{split}
   \end{equation}
   where $y$ is the label of the training image $x$.
   $s$ is a scale factor, $m$ is the margin, and $\theta_{y}=\arccos{\langle\omega^{y}, \phi{(x)}\rangle}$ is the angle between the weight $\omega^{y}$ ($y$-th prototype of the classifier $\omega$) and the feature $\phi(x)$.
   With $\mathcal{L}_{\rm cls}$, the new model can be properly trained to encode discriminative representations for self test.

    According to \cite{Shen_2020_CVPR}, the old classifier (on top of the old backbone model) embeds the characteristic (\textit{i.e.}, class prototypes) of the old latent space, which can be directly leveraged as the valuable supervision in close-set compatible training. 
    However, in the open-set benchmarks of our universal backward-compatible representation learning task, the off-the-shelf old classifier is inapplicable due to the novel new classes.
    Intuitively, to overcome this limitation, we can modify the off-the-shelf old classifier into a pseudo classifier via (1) extracting the features of the new training set by the old model, and (2) using their class centroids as the pseudo classifier weights.
    We denote the pseudo old classifier as $\hat{\omega}_{\rm o}$, and the backward-compatible loss can be formulated as
    \begin{equation}\label{eq:pseudo_cls}
    \begin{split}
        \mathcal{L}_{\rm uniBCT}=\ell_{\rm arc}(\hat{\omega}_{\rm o}, \phi_{\rm n}),
    \end{split}
   \end{equation}
   where $\ell_{\rm arc}$ is the form of ArcFace loss.
   $\mathcal{L}_{\rm uniBCT}$ regularizes to push the new features be closer to their corresponding old class centroids in order to align the old and new latent spaces.
   
    It is notable that the quality of pseudo old prototypes is essential to the training of feature backward compatibility.
    Due to the domain gap (including data gap and category gap) between old and new training sets in open-set model upgrading scenarios, 
    the pseudo old prototypes generated by the simple average operation inevitably carry some noise, affecting the representation learning of backward compatibility. 
    To tackle the challenge, we introduce a novel structural algorithm to refine the prototypes via a fully-connected graph.
  
   \vspace{8pt}
   \noindent \textbf{Structural Prototype Refinement.}  
   As illustrated in Figure \ref{fig:method}, we improve the old prototypes through knowledge propagation under the assumption that visually similar samples of the same class should have close-by old features. 
   We use the training new model to measure their similarities since the new model is expected to have the stronger model capability and could encode more discriminative representations for more accurate similarity measurement.
   
   Specifically, we construct a fully-connected undirected graph $G = (V, E)$ for each class, where $V$ and $E$ represent its vertices and edges.
   In our context, each \textit{old} feature $v_{\rm o}\in\mathbb{R}^d$ serves as a vertex, and features of the same class can be denoted as a matrix $V\in\mathbb{R}^{m\times d}$, where $d$ is the feature dimension and $m$ is the sample number for a certain class.
   The edges among vertices are the similarity scores between pairwise samples, which are measured by cosine similarity, \textit{i.e.}, $\langle v_{\rm n}^i, v_{\rm n}^j\rangle$. Note that we use \textit{new} model features to measure the similarity.
   All the edges of a graph $G$ can be denoted as a symmetric matrix $E$.
   And we further normalize it by row,
      \begin{equation}\label{eq:sim_matrix}
      \begin{split}
      \tilde{E}(i,j) = 
      \begin{cases}
      \frac{{\exp}(\langle v_{\rm n}^i, v_{\rm n}^j\rangle/\tau)}{\sum_{j \neq i}{\exp}(\langle v_{\rm n}^i, v_{\rm n}^j\rangle/\tau)}, \quad &i \neq j \\
      0, \quad &i = j\\
      \end{cases}
      \end{split}
   \end{equation}
   where $\tau$ is the temperature hyper-parameter and 
   a lower temperature leads to a sharper probability distribution.
   
   Each node in the graph randomly visits neighbor images driven by transition probabilities (\ie, similarity scores). Similar nodes (neighbors) are enhanced by each other and closer to the real center of the current class. The outlier features would also be rectified by other nodes.
   Such a propagation process can be formulated as,
   \begin{equation}\label{eq:rw_cls}
       \begin{split}
           V^{(t)} &= \tilde{E}V^{(t-1)}, 
       \end{split}
   \end{equation}
   where $t$ is the iteration times.

   The initial feature matrix $V^{(0)}$ is aggregated to avoid potential collapse in the propagation process, that is,
   \begin{equation}\label{eq:unclosed}
      V^{(t)} = \lambda  \tilde{E}V^{(t-1)}  + (1 - \lambda)V^{(0)},
   \end{equation}
   where $\lambda \in [0, 1]$ is the aggregation weight. 
   When $t$ tends to infinity, Eq.~(\ref{eq:unclosed}) has a converged close form (the proof is provided in our supplemental materials),
   \begin{equation}\label{eq:closed}
      V^{(\infty)} = (1 - \lambda)(I - \lambda \tilde{E})^{-1}V^{(0)},
   \end{equation}
   where $I$ is an identity matrix and $(\cdot)^{-1}$ denotes matrix inverse operation.
   Once $V^{\infty}$ is obtained, the class prototype $\hat{\omega}_{\rm o}$ could be computed by column-wise average pooling of $V^{\infty}$,
   \begin{equation}\label{eq:refined_classifier}
    \hat{\omega}_{\rm o}(j) = \frac{1}{m} \sum_{i=1}^{m} V^{(\infty)}(i, :),
   \end{equation}
   where $\hat{\omega}_{\rm o}(j)\in \mathbb{R}^d$ is the $j$-th pseudo prototype, and $m$ is the number of vertices belonging to the $j$-th class.
   The refined class prototype $\hat{\omega}_{\rm o}$ are used as supervision signals in the universal backward compatible loss $\mathcal{L}_{\rm uniBCT}$ (Eq. (\ref{eq:pseudo_cls})).

   Compared with the vanilla average-based prototype, our introduced structural prototype refinement effectively alleviates the outlier effects by propagating and aggregating the knowledge from neighbor features of the same class.
   
\begin{table*}[!th]
   
   \renewcommand\arraystretch{1.4}
   \begin{center}
   \setlength{\tabcolsep}{2 mm}{
   \footnotesize
      \begin{tabular}{ccccccccccc}
         \toprule
         \multirow{3}{*}{\makecell[c]{Scenarios}} & \multirow{3}{*}{\makecell[c]{Model$_{\rm old}$}} & \multirow{3}{*}{\makecell[c]{Model$_{\rm new}$}} & \multirow{3}{*}{\makecell[c]{Training Set}} & \multirow{3}{*}{\makecell[c]{Comp. Loss}} & \multicolumn{2}{c}{1:1 Verification} & \multicolumn{4}{c}{1:N Identification} \\
         ~  & ~ & ~ & ~ & ~ & Cross Test & Self Test & \multicolumn{2}{c}{Cross Test} & \multicolumn{2}{c}{Self Test} \\
         ~  & ~ & ~ & ~ & ~ & TAR@FAR & TAR@FAR & Top1 & Top5 & Top1 & Top5 \\
         \midrule
          \multirow{6}{*}{\makecell[c]{ \rotatebox{90}{Extended-data} }} & \cellcolor{mygray} $\phi_{\rm o}^{r18}$ & \cellcolor{mygray}~ & \cellcolor{mygray}$D_{\rm o}^{30\%-d}$ & \cellcolor{mygray}- & \cellcolor{mygray}- & \cellcolor{mygray}93.36 & \cellcolor{mygray}- & \cellcolor{mygray}- & \cellcolor{mygray}69.90 & \cellcolor{mygray}75.88 \\
          ~ & \cellcolor{mygray}~ & \cellcolor{mygray}$\phi_{\rm n}^{r18}$ & \cellcolor{mygray}$D_{\rm n}^{100\%}$ & \cellcolor{mygray}- & \cellcolor{mygray}- & \cellcolor{mygray}96.35 & \cellcolor{mygray}- & \cellcolor{mygray}- & \cellcolor{mygray}80.67 & \cellcolor{mygray}85.14\\ 
         ~ & $\phi_{\rm o}^{r18}$ & $\phi_{\rm n}^{r18}$  & $D_{\rm n}^{100\%}$ & $\mathcal{L}_{\rm regress}$ & 0.12 & 94.78 & 8.12 & 10.43 & 76.34 & 80.88 \\
         ~ & $\phi_{\rm o}^{r18}$ & $\phi_{\rm n}^{r18}$ & $D_{\rm n}^{100\%}$ & $\mathcal{L}_{\rm contra}$ & 92.26 & 94.58 & \textbf{73.36} & 81.35 & \textbf{80.90} & \textbf{85.99} \\
         ~ & $\phi_{\rm o}^{r18}$ & $\phi_{\rm n}^{r18}$ & $D_{\rm n}^{100\%}$ & $\mathcal{L}_{\rm uniBCT}^*$ & 93.88 & 94.62 & 72.46 & 81.25 & 80.51 & 84.78 \\
         ~ & $\phi_{\rm o}^{r18}$ & $\phi_{\rm n}^{r18}$ & $D_{\rm n}^{100\%}$ & $\mathcal{L}_{\rm uniBCT}$ & \textbf{94.13} & \textbf{94.85} & 72.89 & \textbf{81.77} & 80.83 & 85.95 \\

         \hline
         \multirow{6}{*}{\makecell[c]{ \rotatebox{90}{Open-data} }}  & \cellcolor{mygray}$\phi_{\rm o}^{r18}$ & \cellcolor{mygray}~ & \cellcolor{mygray}$D_{\rm o}^{30\%-d}$ & \cellcolor{mygray}- & \cellcolor{mygray}- & \cellcolor{mygray}93.36 & \cellcolor{mygray}- & \cellcolor{mygray}- & \cellcolor{mygray}69.90 & \cellcolor{mygray}75.88 \\
         ~ & \cellcolor{mygray}~ & \cellcolor{mygray}$\phi_{\rm n}^{r18}$ & \cellcolor{mygray}$D_{\rm n}^{70\%-d}$ & \cellcolor{mygray}- & \cellcolor{mygray}- & \cellcolor{mygray}94.28 & \cellcolor{mygray}- & \cellcolor{mygray}- & \cellcolor{mygray}75.55 & \cellcolor{mygray}80.24\\ 
         ~ & $\phi_{\rm o}^{r18}$ & $\phi_{\rm n}^{r18}$ & $D_{\rm n}^{70\%-d}$ & $\mathcal{L}_{\rm regress}$ & 0.02 & 94.51 & 7.36 & 9.12 & 73.21 & 78.84 \\
         ~ & $\phi_{\rm o}^{r18}$ & $\phi_{\rm n}^{r18}$ & $D_{\rm n}^{70\%-d}$ & $\mathcal{L}_{\rm contra}$ & 92.23 & 94.42 & 70.34 & 78.20 & 76.69 & 81.75 \\
         ~ & $\phi_{\rm o}^{r18}$ & $\phi_{\rm n}^{r18}$ & $D_{\rm n}^{70\%-d}$ & $\mathcal{L}_{\rm uniBCT}^*$ & 93.75 & 94.37 & 70.35 & 77.68 & 76.54 & 81.69 \\
         ~ & $\phi_{\rm o}^{r18}$ & $\phi_{\rm n}^{r18}$ & $D_{\rm n}^{70\%-d}$ & $\mathcal{L}_{\rm uniBCT}$ & \textbf{94.18} & \textbf{94.52}  & \textbf{71.42} & \textbf{79.14} & \textbf{76.88} & \textbf{81.92} \\

         \hline
         \multirow{6}{*}{\makecell[c]{ \rotatebox{90}{Extended-class} }}  & \cellcolor{mygray}$\phi_{\rm o}^{r18}$ & \cellcolor{mygray}~ & \cellcolor{mygray}$D_{\rm o}^{30\%-c}$ & \cellcolor{mygray}- & \cellcolor{mygray}- & \cellcolor{mygray}92.95 & \cellcolor{mygray}- & \cellcolor{mygray}- & \cellcolor{mygray}68.84 & \cellcolor{mygray}74.72 \\
         ~ & \cellcolor{mygray}~ & \cellcolor{mygray}$\phi_{\rm n}^{r18}$ & \cellcolor{mygray}$D_{\rm n}^{100\%}$ & \cellcolor{mygray}- & \cellcolor{mygray}- & \cellcolor{mygray}96.35 & \cellcolor{mygray}- & \cellcolor{mygray}- & \cellcolor{mygray}80.67 & \cellcolor{mygray}85.14\\ 
         ~& $\phi_{\rm o}^{r18}$ & $\phi_{\rm n}^{r18}$ & $D_{\rm n}^{100\%}$ & $\mathcal{L}_{\rm regress}$ & 0.08 & 93.21 & 7.55 & 9.67 & 74.15 & 78.72 \\
         ~& $\phi_{\rm o}^{r18}$ & $\phi_{\rm n}^{r18}$ & $D_{\rm n}^{100\%}$ & $\mathcal{L}_{\rm contra}$ & 92.70 & 94.53 & 71.83 & \textbf{79.26} & 78.43 & 83.76 \\
         ~ & $\phi_{\rm o}^{r18}$ & $\phi_{\rm n}^{r18}$ & $D_{\rm n}^{100\%}$ & $\mathcal{L}_{\rm uniBCT}^*$ & 93.54 & 94.32 & 71.67 & 79.33 & 78.51 & 84.14 \\
         ~ & $\phi_{\rm o}^{r18}$ & $\phi_{\rm n}^{r18}$ & $D_{\rm n}^{100\%}$ & $\mathcal{L}_{\rm uniBCT}$ & \textbf{93.75} & \textbf{94.55} & \textbf{72.02} & 79.13 & \textbf{78.84} & \textbf{84.33} \\

         \hline
         \multirow{6}{*}{\makecell[c]{ \rotatebox{90}{Open-class} }}  & \cellcolor{mygray}$\phi_{\rm o}^{r18}$ & \cellcolor{mygray}~ & \cellcolor{mygray}$D_{\rm o}^{30\%-c}$ & \cellcolor{mygray}- & \cellcolor{mygray}- & \cellcolor{mygray}92.95 & \cellcolor{mygray}- & \cellcolor{mygray}- & \cellcolor{mygray}68.84 & \cellcolor{mygray}74.72 \\
         ~ & \cellcolor{mygray}~ & \cellcolor{mygray}$\phi_{\rm n}^{r18}$ & \cellcolor{mygray}$D_{\rm n}^{70\%-c}$ & \cellcolor{mygray}- & \cellcolor{mygray}- & \cellcolor{mygray}94.28 & \cellcolor{mygray}- & \cellcolor{mygray}- & \cellcolor{mygray}75.55 & \cellcolor{mygray}80.24\\ 
         ~ & $\phi_{\rm o}^{r18}$ & $\phi_{\rm n}^{r18}$ & $D_{\rm n}^{70\%-c}$ & $\mathcal{L}_{\rm regress}$ & 0.01 & 92.78 & 6.88 & 8.12 & 70.26 & 75.95 \\
         ~ & $\phi_{\rm o}^{r18}$ & $\phi_{\rm n}^{r18}$ & $D_{\rm n}^{70\%-c}$ & $\mathcal{L}_{\rm contra}$ & 92.51 & \textbf{94.24} & 66.51 & 75.82 & 73.63 & 79.96 \\
         ~ & $\phi_{\rm o}^{r18}$ & $\phi_{\rm n}^{r18}$ & $D_{\rm n}^{70\%-c}$ & $\mathcal{L}_{\rm uniBCT}^*$ & 93.35 & 93.96 & 67.14 & 76.38 & 74.21 & 80.28 \\
         ~ & $\phi_{\rm o}^{r18}$ & $\phi_{\rm n}^{r18}$ & $D_{\rm n}^{70\%-c}$ & $\mathcal{L}_{\rm uniBCT}$ & \textbf{93.46} & 94.10 & \textbf{67.47} & \textbf{77.01} & \textbf{74.79} & \textbf{81.22} \\

         \hline
         \multirow{6}{*}{\makecell[c]{ \rotatebox{90}{Identical-data} }} &\cellcolor{mygray} $\phi_{\rm o}^{r18}$ & \cellcolor{mygray}~ & \cellcolor{mygray}$D_{\rm o}^{30\%-d}$ & \cellcolor{mygray}- & \cellcolor{mygray}- & \cellcolor{mygray}93.36 & \cellcolor{mygray}- & \cellcolor{mygray}- & \cellcolor{mygray}69.90 & \cellcolor{mygray}75.88 \\
         ~ & \cellcolor{mygray}~ & \cellcolor{mygray}$\phi_{\rm n}^{r50}$ & \cellcolor{mygray}$D_{\rm n}^{30\%-d}$ & \cellcolor{mygray}- & \cellcolor{mygray}- & \cellcolor{mygray}94.97 & \cellcolor{mygray}- & \cellcolor{mygray}- & \cellcolor{mygray}70.21 & \cellcolor{mygray}76.34\\ 
         ~ & $\phi_{\rm o}^{r18}$ & $\phi_{\rm n}^{r50}$ & $D_{\rm o}^{30\%-d}$ & $\mathcal{L}_{\rm regress}$ & 0.11 & 93.78 & 7.73 & 9.35 & 67.41 & 73.43 \\
         ~ & $\phi_{\rm o}^{r18}$ & $\phi_{\rm n}^{r50}$ & $D_{\rm o}^{30\%-d}$ & $\mathcal{L}_{\rm contra}$ & 92.53 & 95.58 & 64.67 & \textbf{74.69} & 70.49 & 77.13 \\
         ~ & $\phi_{\rm o}^{r18}$ & $\phi_{\rm n}^{r50}$ & $D_{\rm o}^{30\%-d}$ & $\mathcal{L}_{\rm uniBCT}^*$ & 94.40 & 95.42 & 67.38 & 73.25 & 70.57 & 78.34 \\
         ~ & $\phi_{\rm o}^{r18}$ & $\phi_{\rm n}^{r50}$ & $D_{\rm o}^{30\%-d}$ & $\mathcal{L}_{\rm uniBCT}$ & \textbf{94.59} & \textbf{95.63} & \textbf{67.71} & 73.81 & \textbf{70.66} & \textbf{78.76} \\

         \bottomrule
      \end{tabular} 
      }
   \end{center}
   \vspace{-8pt}
   \caption{Comparison of baselines and our proposed approach on IJB-C dataset in universal backward-compatible scenarios, including five different benchmarks. 
   The architectures are ResNet18 (r18) and ResNet50 (r50).
   $\mathcal{L}_{\rm uniBCT}^*$ denotes the \textit{vanilla} version of universal backward-compatible loss where the pseudo prototypes are simply averaged over the raw old features.
   $\mathcal{L}_{\rm uniBCT}$ uses our introduced structural prototype refinement algorithm to improve the pseudo classifier and achieves the optimal performance.
   We evaluate all models in two aspects: (1) For 1:1 verification, the first and second templates are extracted by the new and old model in Cross-Test (CT), and they are processed by the same new model in Self-Test (ST). TAR@FAR=$1e^{-4}$ is adopted as the compatible metric. (2) For 1:N Identification, the query and gallery set are extracted by the new and old models respectively in CT. We report the retrieval accuracy in terms of top1 and top5.
   }
  \vspace{-8pt}
   \label{tab:ijbc}
\end{table*}

   \section{Experiments} 
   \label{sec:experiments}
   To perform a thorough evaluation, we estimate our method (\textit{Uni}BCT) under all compatible settings on the large-scale face recognition dataset. 
   Satisfying results indicates the effectiveness and robustness of our approach.
   
   \subsection{Experimental Setup}
   \noindent \textbf{Datasets.} 
   \textbf{MS-Celeb-1M (MS1M)}~\cite{guo2016ms} is a large-scale face recognition training dataset, which consists of about 10 million images with 1 million identities. Since the original MS1M dataset includes abundant noisy images, we adopt MS1Mv3~\cite{deng2019arcface} as the training set, which is made up of 5,179,510 training images with 93,431 labels.
   \textbf{IJB-C}~\cite{ijbc_maze2018iarpa}, a challenging benchmark, is utilized as the open-set evaluation dataset, which has around 1.3 million images. For verification task, there are 469,376 templates pairs. For identification task, the query set contains 19,593 images and the gallery set consists of 3,531 images.

   \vspace{8pt}
   \noindent \textbf{Metric.} We employ two standard test protocols in face recognition: (1) 1:1 verification calculates the true acceptance rate (TAR) at different false acceptance rates (FAR) for template pairs. In Cross-Test, we extract the first template with the new model, and the second with the old model. (2) 1:N identification evaluates the retrieval accuracy at top-k. In Cross-Test, we process the query set (prob images) and the gallery  set (template images) with the new and old model, respectively.

   \vspace{8pt}
   \noindent \textbf{Training Details.} We use 4 NVIDIA V100 GPUs for training. The training index file is split with fixed random seed 666. We adopt ResNet18 and ResNet50~\cite{he2016deep} architectures as the backbones of the old and new models; one Fully Connected layer is followed to project the output dimension into 512. We adopt standard stochastic gradient descent (SGD) to optimize the model parameters. The learning rate is set to 0.1 and decreases 10 times at the 20th, 26th and 32th epoch. The training stops after 35 epochs. The weight decay is set to $10^{-4}$ and momentum is 0.9. Batch size is set to 256. 
   The scale factor $s$ and margin $m$ in Eq.~\ref{eq:arcface_loss} are 64, 0.5 following the default setting\footnote{\href{https://github.com/deepinsight/insightface}{https://github.com/deepinsight/insightface}}.
   In graph-based prototype refinement, we set $\lambda$ to 0.9, $T$ to 0.05. {The pseudo code of \textit{Uni}BCT can be found in supplemental materials}.

   \subsection{Analysis of \textit{Uni}BCT}

   \noindent \textbf{Effectiveness of Structural Prototype Refinement.}
   Since the quality of pseudo prototypes has essential impact on the backward-compatible learning, we introduce a structural prototype refinery mechanism to improve old features by allocating knowledge from their neighbors. As illustrated in  Table \ref{tab:ijbc}, our method ($\mathcal{L}_{\rm uniBCT}$) not only fulfills the requirement of feature compatible training, but also well boosts the baseline method ($\mathcal{L}_{\rm uniBCT}^*$), which adopts vanilla prototypes for training. 
   
   In addition, an alternative approach for refining pseudo prototype is to discard outlier samples which are far away from the class centroids. 
   Specifically, we filter out the top-10\% data that is away from the mean feature vector in each class, and utilize the rest features to generate the class prototype. 
   As shown in Table \ref{tab:prototype_refinement}, it (denoted as ``drop avg.'') performs worse than the proposed graph-based refinement (denoted as ``refined avg.''). 
   That is because the distribution of the old features is noisy and unreliable, the drop strategy only refers to the old distribution while our structural refinement utilizes the sample similarities in the new latent space as propagation guidance.

   \begin{table}[t]
   \small
      \renewcommand\arraystretch{1.3}
         \begin{center}
         \setlength{\tabcolsep}{0.8 mm}{
            \begin{tabular}{lcccccc}
               \toprule
               \multirow{2}{*}{\makecell[c]{Method}} & \multirow{2}{*}{\makecell[c]{Prototype}}  & \multicolumn{2}{c}{1:1 Verification}  &  \multicolumn{3}{c}{1:N Identification} \\
               ~ & ~ & TAR@FAR & Comp.? & Top1 & Top5 & Comp.? \\
               \midrule
               $\phi_{\rm o}^{r18}$ & - & 93.36 & - & 69.90 & 75.88 & - \\
               \midrule
               Ours & vanilla avg. & 93.88 & \cmark & 72.46 & 81.25 & \cmark \\ 
               Ours & drop avg. & 94.03 & \cmark & 72.35 & 80.97 & \cmark \\
               Ours & refined avg. & 94.13 & \cmark & 72.89 & 81.77 & \cmark \\ 
               \bottomrule
            \end{tabular}
            }
         \end{center}
         \caption{The comparison of different prototypes for the old pseudo classifier. ``Refined avg.'' denotes our optimal solution of structural prototype refinement. The results are reported on IJB-C (extended-data) in terms of 1:1 verification (TAR@FAR=$1e^{-4}$).}
         \label{tab:prototype_refinement}
   \end{table}

    \vspace{8pt}
   \noindent \textbf{Compare to Other Form of Constraints.} 
   The old prototype represents the global contents of the old model,  in the meanwhile, each old feature indicates local details. Therefore, directly maximizing the similarity between the new feature and the corresponding old feature is an alternative choice to achieve compatibility. Specifically, one direct way is to minimize the Euclidean distance between the old and new features extracted from the same image:
   \begin{equation}\label{eq:regression_loss}
      \mathcal{L}_{\rm regress}(\phi_{\rm n}, \phi_{\rm o}) = -\frac{1}{|{\cal D}_{\rm n}|} \sum_{x\in {\cal D}_{\rm n}} \Vert \phi_{\rm n}(x) - \phi_{\rm o}(x) \Vert^2.
   \end{equation}
   As demonstrated in Table \ref{tab:ijbc}, 
   we notice that feature regression fails in all settings. The reason might be that simply minimizing the distance between positive pairs is not enough.
   Thus we turn to another solution, \ie, pulling the new-old positive pairs close and pushing away the negative pairs in the form of contrastive learning.
   Considering each new feature $(\phi_{\rm n}(x^i), y_i)$ as the anchor, the positive set consists of old features with the same class ${\cal P}(i)=\{\phi_{\rm o}(x^j) | y_j= y_i\}$, and the negative set is comprised of the other old features ${\cal N}(i)=\{\phi_{\rm o}(x^j) | y_j\neq y_i\}$. To simplify the training process, we only consider one positive pair $(\phi_{\rm n}(x^i),\phi_{\rm o}(x^i))$. 
   The compatible loss is formulated as,
   \begin{equation}
   \label{eq:contra}
      \begin{split}
      &\mathcal{L}_{\rm contra}(\phi_{\rm n}, \phi_{\rm o})\\ 
      &= -\frac{1}{|{\cal D}_{\rm n}|} \sum_{x^i\in {\cal D}_{\rm n}}{ \log \frac{e^{({{\phi_{\rm n}(x^i)}\cdot \phi_{\rm o}(x^i)/\tau})} }{ \sum_{k\in\{x^i,\mathcal{N}(i)\}}e^{({{\phi_{\rm n}(x^i)} \cdot \phi_{\rm o}(k)/\tau})} } },
      \end{split}
   \end{equation}
   where $\tau$ is a temperature hyper-parameter.
    As shown in Table ~\ref{tab:ijbc}, the performance of \textit{Uni}BCT surpasses the other losses in terms of Cross Test and Self Test.
    Our \textit{Uni}BCT loss adopts the classification-like form following~\cite{Shen_2020_CVPR} and it considers global intra-class and inter-class relations. In contrast, the contrastive loss (Eq. (\ref{eq:contra})) only considers the classes in the current mini-batch, neglecting the global information.

   \vspace{8pt}
   \noindent \textbf{Close-set \vs Open-set.} 
   For 1:1 verification task, our method achieves remarkable performance in all close-set and open-set scenarios.
   For 1:N identification task, the empirical criterion (Eq.~(\ref{eq:empirical_criterion})) is satisfied in most practical settings, except for the most challenging scenario (\ie, open-class), 
   demonstrating that \textit{Uni}BCT can properly alleviate the category gap
   but cannot entirely solve it.
   Even though, we still outperforms other competing methods, indicating the effectiveness of \textit{Uni}BCT.

   \begin{table}[t]
   \small
      \renewcommand\arraystretch{1.3}
         \begin{center}
         \setlength{\tabcolsep}{0.8 mm}{
            \begin{tabular}{lcccccc}
               \toprule
               \multirow{2}{*}{\makecell[c]{Method}} & \multirow{2}{*}{\makecell[c]{Comp. Loss}}  & \multicolumn{2}{c}{1:1 Verification}  &  \multicolumn{3}{c}{1:N Identification} \\
               ~ & ~ & TAR@FAR & Comp.? & Top1 & Top5 & Comp.? \\
               \midrule
               $\phi_{\rm o}^{r18}$ & - & 93.36 & - & 69.90 & 75.88 & - \\
               \midrule
               AML & $\mathcal{L}_{\rm regress}$ & 0.12 & \xmark & 8.12 & 10.43 & \xmark \\
               BCT & $\mathcal{L}_{\rm BCT}$ & 94.01 & \cmark & 72.64 & 81.49 & \cmark \\
               Ours & $\mathcal{L}_{\rm uniBCT}^*$ & 93.88 & \cmark & 72.46 & 81.25 & \cmark \\ 
               Ours & $\mathcal{L}_{\rm uniBCT}$ & 94.13 & \cmark & 72.89 & 81.77 & \cmark \\ 
               \bottomrule
            \end{tabular}
            }
         \end{center}
         \vspace{0pt}
         \caption{Compare to state-of-the-art backward-compatible training methods on IJB-C (extended-data). Only extended-data is evaluated here since BCT is inapplicable for other open-set benchmarks. The results are reported in terms of 1:1 verification (TAR@FAR=$1e^{-4}$).}
         \vspace{-8pt}
         \label{tab:compare_sotas}
   \end{table}

   \subsection{Comparison with State-of-the-arts} 
   To indicate our approach \textit{Uni}BCT can consistently surpasses previous compatible training methods in the conventional close-set benchmarks, we conduct the comparison experiments on the ``extended-data'' setup.
   As shown in Table \ref{tab:compare_sotas}, 
   we compare with BCT~\cite{Shen_2020_CVPR} and AML~\cite{budnik2020asymmetric}. Note that \cite{wang2020unified} and \cite{meng2021learning} are not listed as they require extra network parameters which is not fair.

   AML aims to enlarge the similarity of positive pairs, which is the same as the regression loss in Eq. (\ref{eq:regression_loss}).  AML fails to achieve compatibility in face recognition task though it works well in landmark retrieval in its original paper. Regression loss only focuses on decreasing the distance between positive pairs while ignoring the distance restriction between negative pairs, leading to unsatisfactory performance in fine-grained retrieval tasks, like face recognition. 
    
    As we introduced in the method section, in the close-set setup, the off-the-shelf old classifier can directly serve as the old prototypes according to~\cite{Shen_2020_CVPR}.
   To first investigate the difference between the off-the-shelf old classifier ($\omega_{\rm o}$) and the pseudo classifier ($\hat{\omega}_{\rm o}$), we compare $\mathcal{L}_{\rm BCT}$ and $\mathcal{L}_{\rm uniBCT}^*$ (vanilla avg.) in Table~\ref{tab:compare_sotas}. It is notable that the vanilla prototype achieves comparable performance with minor sacrifice, indicating that the class centers of the pseudo classifier may not be as real as those of the trained classifier. However, with our structural prototype refinement method, \textit{Uni}BCT well surpasses the original BCT, which further demonstrates the effectiveness of our method.

   \vspace{-2pt}
   \section{Conclusion}
   \label{sec:conclusion}
   We for the first time introduce the task of universal backward-compatible representation learning, which covers both close-set and open-set compatible training scenarios for real-world model upgrades.
   To tackle the challenge of noisy old prototype features, we propose a simple yet effective method, namely \textit{Uni}BCT, to properly refine the prototypes by propagating and aggregating their neighbors' knowledge. 
   \textit{Uni}BCT trains the new models to encode discriminative and compatible representations in five different benchmarks in a unified manner.
   It is the first step towards universal compatible feature learning, and there's still a long way to go for totally solving this problem. Further studies are called for.
   
\bibliographystyle{named}
\bibliography{ijcai22}

\newpage
\onecolumn
\section*{Appendix}

\paragraph{Proof of Equation (9)}
    \begin{equation}
        \begin{aligned}
        V^{(t)} &= \lambda  \tilde{E}V^{(t-1)}  + (1 - \lambda)V^{(0)} \\
         &= \lambda \tilde{E}(\lambda \tilde{E}V^{(t-2)}+(1-\lambda)V^{(0)}) +(1-\lambda)V^{(0)} \\
         &= \lambda^2 \tilde{E}^2 V^{(t-2)} + \lambda (1-\lambda)\tilde{E}V^{(0)} + (1-\lambda)V^{(0)} \\
         &=  \lambda^3 \tilde{E}^3 V^{(t-3)} + \lambda^2 (1-\lambda) \tilde{E}^2 V^{(0)} + \lambda (1-\lambda) \tilde{E} V^{(0)} + (1-\lambda)V^{(0)} \\
         &=\cdots \\
         &=\lambda^t \tilde{E}^t V^{(0)} + \sum_{i=1}^{t-1}{\lambda^i(1-\lambda)\tilde{E}^i V^{(0)}} + \lambda^0(1-\lambda)\tilde{E}^0 V^{(0)} \\
         &= \lambda^t\tilde{E}^t V^{(0)} + \sum_{i=0}^{t-1}{\lambda^i(1-\lambda)\tilde{E}^i V^{(0)}}
        \end{aligned}
    \end{equation} 
 
    when $t$ tends to infinity, $\lambda^t=0 (\lambda<1)$, and we can obtain the following formula,
    \begin{equation}
        \begin{aligned}
          V^{(\infty)} = (1-\lambda) (\sum_{i=0}^{\infty}{\lambda^i \tilde{E}^i})V^{(0)} = (1-\lambda) (I - \lambda \tilde{E})^{-1} V^{(0)}.
        \end{aligned}
    \end{equation}

\paragraph{Pseudo Training Code}
   \label{sec:conclusion}
   For reproducibility, we illustrate the core algorithm of \textit{Uni}BCT in Alg.~\ref{alg:uni_bct} and Alg.~\ref{alg:prototype_generation}.
   For warming up, we first train the new model 10 epochs only with the classification loss ($\eta=0$). In the next 25 epochs, we add the \textit{Uni}BCT loss with $\eta=1$. We generate and refine the old prototype every 10 epochs.

\begin{algorithm*}[h]
\caption{Pseudocode of Universal Backward-Compatible Training (\textit{Uni}BCT) in a PyTorch-like style.}
\label{alg:uni_bct}
\definecolor{codeblue}{rgb}{0.25,0.5,0.5}
\lstset{
  backgroundcolor=\color{white},
  basicstyle=\fontsize{8pt}{8pt}\ttfamily\selectfont,
  columns=fullflexible,
  breaklines=true,
  captionpos=b,
  commentstyle=\fontsize{8pt}{8pt}\color{codeblue},
  keywordstyle=\fontsize{8pt}{8pt},
}
\begin{lstlisting}[language=python,escapeinside={(*}{*)}]
# old_model: pretrained and fixed old encoder, no gradient
# new_model: new encoder, new_model.fc is the classifier

for epoch in range(35):
    if epoch in [10,20]: 
        # generate the old prototype with structural refinery mechanism
        with torch.no_grad():
            w_o_hat = PrototypeGeneration(new_model, old_model, loader)
    new_model.train()
    for (x, labels) in loader:  # load a mini-batch x
        new_feat = new_model.forward(x)
        
        # Classification loss, Eqn.(4)
        cls_logits = new_model.fc(new_feat)
        loss = ArcFaceLoss(cls_logits, labels)
        
        if epoch>10:
            # universal backward-compatible training loss, Eqn.(5)
            uni_bct_loss = UniBCTLoss(w_o_hat, new_feat)
            loss = cls_loss + uni_comp_loss
    
        loss.backward()
        update(new_model.params)
\end{lstlisting}
\end{algorithm*}

\begin{algorithm*}[h]
\caption{Pseudocode of Prototype Generation and Refinement in a PyTorch-like style.}
\label{alg:prototype_generation}
\definecolor{codeblue}{rgb}{0.25,0.5,0.5}
\lstset{
  backgroundcolor=\color{white},
  basicstyle=\fontsize{8pt}{8pt}\ttfamily\selectfont,
  columns=fullflexible,
  breaklines=true,
  captionpos=b,
  commentstyle=\fontsize{8pt}{8pt}\color{codeblue},
  keywordstyle=\fontsize{8pt}{8pt},
}
\begin{lstlisting}[language=python,escapeinside={(*}{*)}]
# cls_num: the number of classes

def PrototypeGeneration(new_model, old_model, loader):
    with torch.no_grad():
        for (x, labels) in loader:  # extract old and new features
            new_feat = new_model.forward(x)
            old_feat = old_model.forward(x)
            
        new_feat_list = [[] for _ in range(cls_num)]
        old_feat_list = [[] for _ in range(cls_num)]
        old_prototype = zeros(cls_num,)
        for i, label in enumerate(labels):  # aggregate by category
            new_feat_list[label].append(new_feat[i,:].unsqueeze_(0))
            old_feat_list[label].append(old_feat[i,:].unsqueeze_(0))
        for label in labels:
            old_vertices = torch.stack(old_feat_list[label])
            new_vertices = torch.stack(new_feat_list[label])
            edges = torch.mm(new_vertices, new_vertices.t())
            identity = torch.eye(edges.size(0))
            mask = torch.eye(edges.size(0), edges.size(0)).bool()
            edges.masked_fill_(mask, -1e9))
            edges = softmax(edges, dim=0)
            # Eq. (9)
            edges = (1-lambda)*torch.inverse(identity - lambda * edges)
            old_vertices = torch.mm(edges, old_vertices)
            # Eq. (10)
            old_prototype[label] = old_vertices.mean(dim=0)
    return old_prototype
        
\end{lstlisting}
\end{algorithm*}

\end{document}